\documentclass[11pt]{article}
\usepackage[a4paper, margin=1.0in]{geometry}
\usepackage{ragged2e}
\usepackage{tcolorbox}   

\usepackage[utf8]{inputenc}

\usepackage[style=apa,sortcites=true,sorting=nyt,backend=biber]{biblatex}
\usepackage[colorlinks=true,linkcolor=blue,citecolor=blue,urlcolor=blue]{hyperref}
\bibliography{refs}

\usepackage{placeins}

\usepackage{nameref,hyperref}

\usepackage[nopatch=footnote]{microtype}
\DisableLigatures[f]{encoding = *, family = * }

\justifying
\usepackage{setspace} 
\usepackage{changepage}

\usepackage[aboveskip=1pt,labelfont=bf,labelsep=period,singlelinecheck=off]{caption}

\makeatletter
\renewcommand{\@biblabel}[1]{\quad#1.}
\makeatother

\usepackage{lastpage,fancyhdr,graphicx}
\usepackage{epstopdf}
\usepackage{color}

\usepackage{glossaries}

\definecolor{Gray}{gray}{.25}

\usepackage{graphicx}

\usepackage{sidecap}

\usepackage{wrapfig}
\usepackage[pscoord]{eso-pic}
\usepackage[fulladjust]{marginnote}
\reversemarginpar

\makeglossaries
\newglossaryentry{predictive coding}
{
    name=predictive coding,
    description={Predictive coding is a unifying theory that has emerged in computational neuroscience, postulating that the brain performs inference on hierarchical predictive models of the environment through the computation of local prediction errors.}
}

\newglossaryentry{mean-field approximation}
{
    name=mean-field approximation,
    description={The mean-field approximation is a technique that can be used to calculate approximate partition functions for systems composed of interacting particles. When applied to hierarchical probabilistic models, it allows to approximate the value of a node using the information available in the most proximal nodes, without computing the dependencies for the entire system, which is often intractable.}
}
\newglossaryentry{hierarchical gaussian filter}
{
    name=hierarchical Gaussian filter,
    description={Hierarchical Gaussian filters (HGFs) are a widely used family of models in computational psychiatry that unifies reinforcement learning and Bayesian inference. While it can be described as a recurrent neural network, it is often more conveniently described using the generative model it inverts. When applied straightforwardly to time series, the hierarchical Gaussian filters act as Bayesian filters that update their expectations using errors of predictions from the previous time point, an approach that is similar to both the Kalman filter (which is a special case of an HGF) and Rescorla-Wagner update rules. In HGFs, the learning rate is not fixed but influenced by higher levels of the hierarchy, generally encoding the volatility of the environment. Recent developments have turned this model family into a generic framework that can handle arbitrarily sized networks of distributions.}
}

\begin{document}

\begingroup
\centering
\Huge\bfseries
pyhgf: A neural network library for predictive coding\par
\endgroup

\bigskip

\begin{flushleft}
Nicolas Legrand\textsuperscript{1}\footnote{Corresponding author: \href{mailto:nicolas.legrand@cas.au.dk}{nicolas.legrand@cas.au.dk}},
Lilian Weber\textsuperscript{2},
Peter Thestrup Waade\textsuperscript{1},
Anna Hedvig Møller Daugaard\textsuperscript{1},
Mojtaba Khodadadi\textsuperscript{3},
Nace Miku\v{s}\textsuperscript{1,4},
Christoph Mathys\textsuperscript{1}
\\
\bigskip
\small 1 - Interacting Minds Centre, Aarhus University, Denmark
\\
\small 2 - Department of Psychiatry, University of Oxford, United Kingdom
\\
\small 3 - Scuola Internazionale Superiore di Studi Avanzati (SISSA), Trieste, Italy
\\
\small 4 - Department of Cognition, Emotion, and Methods in Psychology, Faculty of Psychology, University of Vienna, Austria
\\

\end{flushleft}

\section*{Abstract}
\begin{tcolorbox}[colback=gray!10, colframe=black, 
                  arc=1mm, 
                  boxrule=0.0mm, 
                  colframe=white
                  ]

\small{
Bayesian models of cognition have gained considerable traction in computational neuroscience and psychiatry. Their scopes are now expected to expand rapidly to artificial intelligence, providing general inference frameworks to support embodied, adaptable, and energy-efficient autonomous agents. A central theory in this domain is predictive coding, which posits that learning and behaviour are driven by hierarchical probabilistic inferences about the causes of sensory inputs. Biological realism constrains these networks to rely on simple local computations in the form of precision-weighted predictions and prediction errors. This can make this framework highly efficient, but its implementation comes with unique challenges on the software development side. Embedding such models in standard neural network libraries often becomes limiting, as these libraries' compilation and differentiation backends can force a conceptual separation between optimization algorithms and the systems being optimized. This critically departs from other biological principles such as self-monitoring, self-organisation, cellular growth and functional plasticity. In this paper, we introduce \texttt{pyhgf}: a Python package backed by JAX and Rust for creating, manipulating and sampling dynamic networks for predictive coding. We improve over other frameworks by enclosing the network components as transparent, modular, and malleable variables in the message-passing steps. The resulting graphs can implement arbitrary algorithms as belief propagation. Moreover, the transparency of core variables can also translate into inference processes that leverage self-organisation principles and express structure learning, meta-learning, or causal discovery as the consequence of network structural adaptation to surprising inputs. The main functions of the library are differentiable and seamlessly integrate into sampling or optimization workflows. Additionally, we offer generalized Bayesian filtering and the hierarchical Gaussian filter as key examples of dynamic networks implemented in our library. The source code, tutorials and documentation are hosted under the main repository at \url{https://github.com/ComputationalPsychiatry/pyhgf}.} \\

\end{tcolorbox}

\textbf{keywords}: \textit{predictive coding, hierarchical Gaussian filter, computational psychiatry, Bayesian networks, neural networks, reinforcement learning}

\clearpage

\section{Introduction}

Bayesian models of cognition describe perception and behaviours as probabilistic inference over the cause of sensory inputs \parencite{Ji:2023}. Modelling these processes at scale to infer computational parameters from human behaviours \parencite{Huys2016, Sandhu2023, Friston2022}, or to implement them into artificial agents \parencite{Mathys2020, DaCosta2022}, is currently a challenge that brings together computational neuroscience and artificial intelligence. However, when considering living organisms, the complexity of models increases and inference becomes especially challenging. While certain inferential processes can sometimes be straightforwardly described and implemented using closed-form solutions, intractability emerges rapidly with models that incorporate multiple information streams, continuous inputs, or hierarchical dependencies common in biological systems. \Gls{predictive coding} \parencite{Rao:1999, Friston:2005} has posited that such complex generative probabilistic models are biologically implemented as hierarchical networks of nodes (i.e. neurons or populations of neurons) that perform simple computations such as message-passing, error signalling and belief propagation in interaction with other proximal units in the hierarchy \parencite{Friston2008, Mikulasch2023, Ororbia2022}. This mechanism could represent a simpler, faster, and energy-efficient alternative to other optimization methods such as backpropagation \parencite{Rumelhart1986}, which is commonly used with artificial neural networks \parencite{Millidge:2022}, or MCMC sampling \parencite{Betancourt2017} in the case of Bayesian inference.

One limiting factor to the wider application of predictive coding neural networks to more complex probabilistic models is the absence of easily accessible open-source toolboxes compatible with modern probabilistic programming and neural network libraries. It is therefore critical for the field to develop a framework that facilitates the implementation of predictive coding models in a manner analogous to how TensorFlow \parencite{tensorflow2015} and PyTorch \parencite{pytorch} have supported the development of conventional neural networks over the past decade. However, this requirement comes with considerable technical challenges on the software development side. This becomes apparent when we consider the divergence between the models that standard neural network libraries are tailored to develop and the dynamic and flexible nature of the models that computational neuroscience and computational psychiatry seek to develop. Biological organisms can implement learning and flexible behaviours not only by adjusting the inner representation of certain quantities but also by leveraging self-monitoring, self-organisation, cellular growth and functional plasticity. Even abstracting from the specific optimisation algorithm or inference method that we aim to apply, few of these features could be implemented in classical deep learning libraries. First, those libraries need to compile to low-level programming languages while maintaining the possibility of automatic differentiation. This often comes at the cost of a restriction of dynamic manipulation of inner variables at execution time. For example, conventional neural networks rely heavily on linear algebra functionalities that require static matrix shapes, which has been a limiting factor for graph neural networks for example (see however how \parencite{PyTorchGeometric} and \parencite{jraph2020} circumvent parts of this problem). Secondly, these frameworks tend to disentangle the optimisation process from the optimized system. While the network is defined through a set of variables only partially transparent, tuning the network relies on the execution of scripts whose steps are hidden from the network, preventing it from reasoning about inference itself. It is therefore crucial for predictive coding, and to adhere to biological realism, that self-monitoring and self-organisation principles could be instantiated more easily. 

The second important limiting factor to a broader application of predictive coding neural networks in reinforcement learning studies and computational psychiatry, in particular, is the possibility to "\textit{observe the observer}" \parencite{Daunizeau2010} and thereby to infer parameters of implied networks from observed decisions and behaviours. This implies that a framework capable of this should expose the resulting networks to optimization and sampling packages for inference on large datasets, for example for multilevel experimental designs like repeated measures or group comparisons. However, performing inference over a set of parameters to inform learning profiles from behaviours requires another inversion of the model and the use of inference techniques like Hamiltonian MCMC sampling \parencite{Betancourt2017}, which require automatic differentiation of the likelihood function, which is not possible out-of-the-box in several programming languages.

Finally, one critical component that goes beyond the concrete implementation and flexibility of such networks is the availability of validated methods and models that can be deployed and adapted into experimental applications. Predictive coding frameworks are available in many flavours \parencite{millidge:predictivecodingreview} and there is a need for robust implementation of those frameworks so that they are easily accessible to non-expert users. The \textit{\gls{hierarchical gaussian filter}} (HGF) \parencite{2011:mathys, 2014:mathys} is a popular inversion scheme for predictive coding-inspired models. Over the past decade, it has been widely adopted in computational psychiatry and reinforcement learning to emulate Bayesian belief updating in agents that are facing changing environments. In this framework, the networks encode the agent's probabilistic inference about latent states of the environment, which are updated in real time by new observations. By encapsulating a hierarchy of Gaussian probability densities, this framework is well-suited for modelling how belief expectations and precisions are propagated in the graph and how they affect value updating \parencite{Sandhu2023}. Many complex cognitive phenomena (e.g. hallucinations and delusions) and psychiatric conditions (e.g. anxiety, autism, schizophrenia), can efficiently be described in this way through altered uncertainty or precision processing at various levels of the hierarchy \parencite{Corlett2019, Reed2020, Powers2017, Lawson2017}. An important factor in the popularity of the HGF in computational neuroscience has been the availability of a Matlab toolbox \parencite{Frssle2021}, together with its documentation and a forum for community support (\url{https://github.com/ComputationalPsychiatry/hgf-toolbox}). This toolbox implements the core components of the framework for experimental neuroscience (i.e., the two-level and three-level binary and continuous HGF along with several variations thereof, and an array of response functions). However, generalisation of the model to arbitrarily sized networks is not provided and the programming language makes it difficult to interface with other Bayesian modelling and neural networks tools.

In this paper, we introduce \texttt{pyhgf}, a neural network library for creating, manipulating and sampling dynamic neural networks for predictive coding. In \texttt{pyhgf}, each local computation is an in-place function operating on the network itself, defined by its attributes, edges, transformations and propagation dynamics. All network components are modular and transparent during propagation, which means that they can be part of the inference process. It natively supports the implementation of the \textit{generalized hierarchical Gaussian filter} (gHGF) \parencite{weber2023}, a fully nodalised neural network structure where belief nodes can be flexibly added or removed without any additional derivations. This step considerably extends the complexity of the networks that are supported without requiring additional work from the user and only involves local computations of prediction, prediction error and posterior updates, as per predictive coding standards. \texttt{pyhgf} is written on top of JAX \parencite{jax2018github}, an XLA and autograd tensor library for Python that supports parallelisation on GPUs and TPUs, as well as in Rust \parencite{rust}, a general-purpose programming language designed for performance, safety, and concurrency. The user can decide which of these two backends to use depending on the type of application. This feature allows flexible and computationally efficient network representation, together with smooth integration with other optimization libraries in the ecosystem \parencite{deepmind2020jax}, both for Bayesian inference (e.g. to iterate HGF models as part of multilevel Bayesian networks) or to interface with other reinforcement learning and neural network libraries \parencite{flax2020github}.

This paper is organised as follows: we first describe dynamic neural networks from a theoretical and programming point of view, with a focus on the generalised hierarchical Gaussian filter for predictive coding, which is a specific instance of such a network. We introduce the proposed framework and highlight key differences both with previous versions \parencite{2011:mathys, 2014:mathys} and other software implementations \parencite{Frssle2021}. In the results section, we illustrate the standard workflow supported by the toolbox, from network development to observing the observer. We implement the classical three-level Hierarchical Gaussian Filter for binary inputs and demonstrate forward fitting, multiple response models, Bayesian inference, multilevel hierarchical modelling, parameter recovery and model comparison. Finally, we discuss how the proposed tool could facilitate the creation and simulation of autonomous agents that dynamically approximate high-dimensional distributions to navigate their environment, how this could streamline the development of active inference and highlight new research lines at the interface between computational neuroscience and artificial intelligence.
\section{Design and Implementation}

\texttt{pyhgf} is a Python library for the creation, manipulation, and inference over dynamic neural networks for predictive coding with a focus on the generalized Hierarchical Gaussian Filter \parencite{2011:mathys, 2014:mathys, weber2023}. Models and theories that imply such networks are becoming ubiquitous in computational neuroscience, and researchers interested in fitting behavioural data to these models require the flexibility of a regular neural network library together with the modularity of a probabilistic framework to perform inference on parameters of interest. In the \texttt{pyhgf} package, we provide the user with an API that provides methods for smoothly interacting with the following two levels of modelling: 

\begin{enumerate}
\item A set of core methods to define, manipulate and update dynamic neural networks for predictive coding. These networks need to provide unique flexibility in their design, which is enabled by giving the user control over a limited set of parameters, accessible both to the user in the design process and to the agent in real-time adaptive behaviours.

\item Higher-level classes for embedding any of these networks as custom likelihood functions in multi-level Bayesian models, or as loss functions in other optimisation libraries. Those classes include fully defined probabilistic distributions that integrate with PyMC \parencite{pymc2023} and tools to help diagnose inference, visualization, and model comparison \parencite{Kumar2019}.
\end{enumerate}

By using these interfaces, the user is able to customize the computational structure of artificial agents to fit a broad range of applications, both in experimental cognitive neuroscience and artificial intelligence. Here, we start by reviewing the core principles on which dynamic neural networks are built in \texttt{pyhgf}, and how this differs from other network libraries.

\subsection{Computational framework}

The design and software implementation of dynamic neural networks for predictive coding has been shaped by a set of requirements. These networks are made of nodes that can store any number of variables. Some variables might be found in other nodes as well, and some might be unique. Nodes are connected with each other through directed edges, and there can exist multiple types of connections in the same graph, denoting different interactions between the nodes. Computation steps in the graph typically occur locally between adjacent nodes for prediction and prediction errors. Multiple types of computation can be defined. The computation steps can be triggered either reactively, by observing events in the surroundings of a node and reacting to them, or they can be scheduled by pre-allocating a sequence of steps that will propagate the information through the graph. Finally, all these components should be transparent to the network itself when performing a given computation, allowing it to meet the demands of self-monitoring and self-organisation principles.

To observe the set of constraints above, computations should follow a strict functional programming framework, meaning that they should be in-place programmatically pure functions operating on the components of the network. Functional programming is natively supported in Rust and also enforced by JAX \parencite{jax2018github} to leverage just-in-time (JIT) compilation and automatic differentiation, therefore departing from object-oriented programming (the definition of classes populated with attributes and methods) that is a central feature of Python. This comes with limitations in the way toolboxes' API can be developed (see for example how this can be handled in \parencite{Kidger2021}). To fully meet the dynamic aspects mentioned previously, the update functions should ideally receive and return all of the following components defining a network:

\begin{enumerate}
\item A list of attributes - the attributes are dictionaries of parameters of a given node
\item A set of lists of edges - the edges are the directed connections between the nodes. All the possible edge types are grouped into a set.
\item A set of update functions. Each function defines a computation and can be parametrised by the index of the triggering node. Possible computations are for example prediction, posterior update or prediction error between two nodes.
\item Unless using a reactive computation scheme, the function should also have access to an ordered sequence of update functions that apply to individual nodes.
\end{enumerate}

\begin{figure}[ht]
\includegraphics[width=\linewidth]{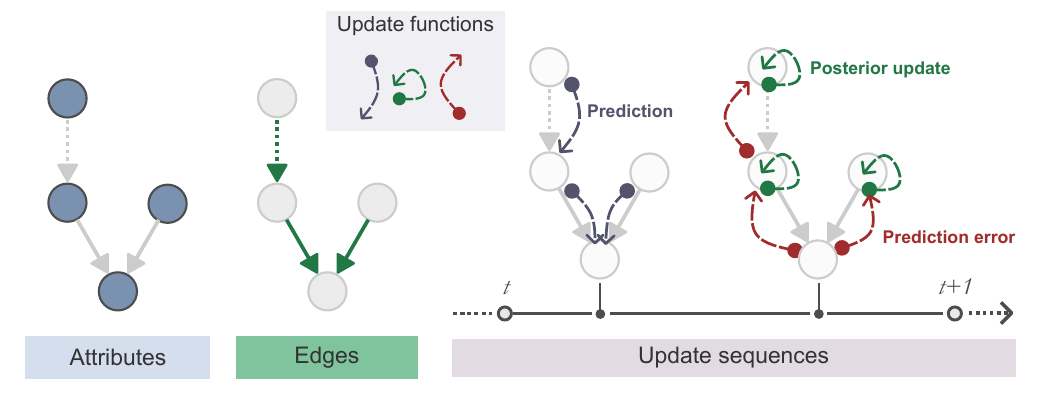}
\captionsetup{font=footnotesize}
\caption{
\color{Gray}
\textbf{The four components of a dynamic network for predictive coding.} \texttt{pyhgf} represents any dynamic network using the combination of four variables: attributes, edges, update functions and update sequences. This modularity allows dissociating update steps and connectivity structures and makes these variables part of the inference process. The creation of a network is read from left to right: \textbf{1. Attributes.} Nodes in the network contain parameters (e.g., sufficient statistics about probability distributions and coupling weights). \textbf{2. Edges.} Nodes can have multiple connection types with each other (e.g., value and volatility coupling). The network's structure is represented in an $m$-dimensional adjacency list that encodes the directed connections with other nodes. Here, dotted and filled lines represent different types of connectivity. \textbf{3. Updates.} Update functions are deterministic transformations operating locally that can access and modify the four sets of variables at run time. \textbf{4. Update sequences.} The update sequence shapes belief propagation. It defines the order in which nodes should be updated when a new observation is presented to the input node(s). By default, prediction propagates from the leaves to the root of the network, while the interleaved sequence of prediction errors and posterior updates follow the inverted path. Updates can also be triggered reactively in which case the propagation starts with the activation of a proximal node.}
\label{fig:networks}
\end{figure}

By defining these four components, and by creating functions that can receive and return all of them, the user can generate arbitrarily sized and structured dynamic neural networks for predictive coding (see \ref{fig:networks}). The first two items define what is usually called a graph, with the addition that it can be directed and multilayered. The last two items shape what is central to predictive coding: the schedule or reactive nature of the propagation of information through the network. Because all of these components are transparent during message-passing computations, learning algorithms can be developed to act on them as a way of inference. Acting on the attributes corresponds to standard inference or other learning algorithms like reinforcement learning. Acting on the size of the networks is comparable to structure learning. Acting on the edges can relate to causal inference, and acting on the update functions or their sequence can implement principles from meta-learning (see \ref{Availability} for more details on possible applications). The hard constraint on transparency of the network component during message passing makes this framework difficult to implement in other graph/neural network libraries (e.g. see \cite{jraph2020, PyTorchGeometric}). Most of these constraints can be met by using a pure JAX implementation \parencite{jax2018github} while remaining compatible with transformations like JIT and automatic differentiations. However, some advanced use cases of dynamic reshaping and edge manipulation might result in degraded performance or incompatibility with certain transformations. When using Rust \parencite{rust} as the backend, all constraints can be met with no cost in terms of performance.

A dynamic network, as implemented in \texttt{pyhgf} is thus defined as the combination of four variables (see \ref{fig:networks}). Let for example $\mathcal{N}_{k}$ be a neural network with $K$ nodes. This network handles in a tuple four parameters of interest:

\begin{equation}
\mathcal{N}_{K} = \{\Theta, \Xi, \mathcal{F}, \Sigma \}
\end{equation}

The variable $\Theta = \{\theta_1, ..., \theta_{k}\}$ represents the nodes' attributes. Attributes can be used to register local information like the sufficient statistics of a probability distribution as well as the coupling weights with other nodes. This variable can also be arbitrarily extended to include other fixed parameters or results from other update steps. In a convolutional neural network, $\Theta$ would for example encode the activation strength. Most standard learning models optimize attributes that belong to this parameter space. 

The second key parameter, tightly linked to the first one, is the adjacency list $\Xi = \{\xi_1, ..., \xi_{k} \}$ that controls the shape of the network. Each item in this set registers the directed connection between node $k$ and other nodes. Networks that exhibit different connectivity structures propagate information differently. The set of directed connections can be multivariate (a node can have different types of connection with other nodes), such as in multilayer networks \parencite{DeDomenico2023}. For example, the nodalised Hierarchical Gaussian Filter \parencite{weber2023} assumes two kinds of coupling between nodes: value and volatility coupling. Every edge $\xi_k$, therefore, contains $m = 2$ sets of node indices in this case, $m$ being the adjacency dimension. By comparison, in a standard recurrent neural network, this variable would define the shape of the layers and their connectivity. Critically, in the proposed framework, this variable is transparent to the update function and can be subject to inference and updates.

The third central component is the set of $n$ in-place update functions $\mathcal{F} = \{f_1, ..., f_n\}$ defining a message passing step operating on the network's parameter set such as:

\begin{equation}
f_n^k(\mathcal{N}_{K}) = \mathcal{N}_{K}'
\end{equation}

In a convolutional neural network, this set of functions would include the linear product of input and weights as well as the activation functions. In the generalised HGF \parencite{weber2023}, this includes three kinds of steps: a prediction (based on previous values and any parent nodes), an update step (based on input from child nodes and the prediction), and the computation of a prediction error. The specific computations in each case depend on the type of coupling between parent and children nodes. Note that the function is parametrized by a target node $k$ to which it applies. This allows for defined local computation where only information from a subset of adjacent nodes is used, such as in particular mean field approximations.

Finally, a fourth component is introduced to control the scheduling of these update steps over time as $\Sigma = [f_1^{n_1}, ..., f_i^{n_k}, f_i \in \mathcal{F}, n \in 1, ..., k ]$. This ordered list describes a sequence of functions parametrized on individual nodes. The update order shapes belief propagation. This component is rarely expressed in the form of a parameter in most conventional applications of neural networks, as well as in the previous implementation of the HGF. This sequence is instead scripted outside the network's closure and, therefore, not accessible during inference or optimization. In the case of predictive coding neural networks, however, finer control over belief propagation might be requested by the user, of a kind which also offers flexibility in the modification of belief propagation dynamically. When using the HGF as implemented in \texttt{pyhgf}, this scheduling can be generated at runtime from the network structure $\Xi$, assuming an ideal belief propagation pattern with a cascade of prediction from the leaves to the roots of the network, and another cascade of prediction error / posterior update pair from the roots to the leaves.

The proposed framework is intended to provide the minimal layout required to create dynamic neural networks for predictive coding. It allows users to create and manipulate the scheduling of updates through a network of nodes while keeping the four components of the networks available for inference. Contrary to other neural networks that rely on matrix multiplication for learning, our networks implement local computations that are run sequentially to propagate beliefs along connectivity paths. This also offers a clear dissociation between components that can be developed separately. It is for example possible to create alternative message-passing algorithms without having to develop an entire library to simulate the networks, and it is possible to implement existing predictive coding frameworks so users can easily apply them to behavioural data. \texttt{pyhgf} natively support the generalised hierarchical Gaussian filter \parencite{weber2023}, a recent development of the Hierarchical Gaussian Filter \parencite{2011:mathys, 2014:mathys} into a nodalised version for predictive coding. In this framework, for example, every node in the network represents a probability distribution of a belief about a latent space in the environment. Beliefs are updated through precision-weighted prediction errors coming from nodes in a lower level of the hierarchy and propagated to higher-level nodes. The exact update functions have been derived in their closed form and can work with arbitrary network architectures \parencite{weber2023}, which makes this model an excellent application of dynamic neural networks as described here. The relatively widespread use of the hierarchical Gaussian filter in computational psychiatry, and the need for advanced Bayesian modelling tools around it, are also good opportunities to extend the original Matlab toolbox \parencite{Frssle2021} by enhancing the modularity and extensibility of the library.

\subsection{Optimization and inference}

While predictive coding itself originates from fields related to signal processing and information theory \parencite{Oliver1952}, the use of predictive coding as a framework for hierarchical inference \parencite{Rao:1999, Friston:2005} in biological neural networks makes it especially well-suited to fields related to experimental neuroscience and computational psychiatry. In this context, the neural networks are components of a cognitive model of the subject on which the experimenter performs inference (i.e., observing the observer \parencite{Daunizeau2010}). For example, in the context of the generalised hierarchical Gaussian filter, the user might be interested in inferring the posterior distribution of tonic volatility at different levels of the hierarchy from observed behaviours.

This kind of reverse inference requires the use of techniques like MCMC sampling or gradient descent which involves the evaluation of several instances of a network, as well as the gradient at evaluation, to find parameters maximizing likelihood. In the Matlab HGF toolbox \parencite{2011:mathys, 2014:mathys}, the inference step is implemented using a variant of the BFGS algorithm, which can be difficult to apply in the context of multilevel models, where there is a particularly pressing need for both high performance and the benefits of automatic differentiation. The \texttt{pyhgf} codebase is entirely written in Python and, as of version 0.2.0, can use JAX \parencite{jax2018github} as a computational backend which can easily deploy code on CPU, GPU and TPU. JAX offers a rapidly growing ecosystem for machine learning \parencite{deepmind2020jax} and artificial intelligence that already includes toolboxes that are conceptually related to predictive coding and Hierarchical Gaussian Filters, such as state-space modelling (\url{https://github.com/probml/dynamax}), reinforcement learning \parencite{hoffman2020}, neural networks \parencite{Kidger2021, flax2020github} or graph neural networks \parencite{jraph2020}. We leverage the automatic differentiation and just-in-time compilation offered by JAX \parencite{jax2018github} to let the networks interface smoothly with other optimization and inference libraries like PyMC \parencite{pymc2023} that support a large range of sampling or variational methods, including Hamiltonian Monte-Carlo methods such as the No-U-Turn Sampler (NUTS) \parencite{Matthew:2014}, an approach that has proved to be highly efficient when scaling to high-dimensional problems. While dimensionality was not a major concern for individual model fittings, this can become problematic if we want to model group-level parameters, and therefore estimate a large number of networks together with hyperpriors (multilevel modelling). Assessing group-level estimates is a crucial step for studies in computational psychiatry, where gaining insights into computational parameters at the population level can inform further diagnosis and classification. In \texttt{pyhgf}, it is possible to apply multilevel modelling to any dynamic neural network handled by the toolbox. In the next section, we will focus on the development workflow using the standard three-level Hierarchical Gaussian Filter as an example.

\FloatBarrier
\section{Examples} \label{Results}

Users interested in using the pyhgf package are referred to the main documentation under the following link: \url{https://github.com/ComputationalPsychiatry/pyhgf}. The documentation features an up-to-date theoretical introduction, API descriptions, and an extensive list of tutorials and examples of various use cases. In this section, we are concerned with the standard analytic workflow of the package: we will explain how to create and manipulate dynamic networks, how to fit the network to a sequence of observations, and how to perform inference and optimization over parameters (see \ref{fig:inference}). This concrete use case will demonstrate how dynamic networks are especially well suited to create modular efficient structures for signal processing and real-time decision-making. These generic systems have direct applications in artificial intelligence and computational neuroscience, focusing on psychiatry, for the robust estimation of learning parameters through multilevel models, model comparison and parameter recovery. These procedures are illustrated in \ref{fig:recovery} using a large simulated dataset.

For the results reported in \ref{fig:inference} and in \ref{fig:recovery}, we used binary observations and binary decision responses (hereafter denoted $u$ and $y$, respectively) from a behavioural experiment using an associative learning task \parencite{Iglesias2021}. Such a dataset is especially well suited for a binary hierarchical Gaussian filter. Here we used the three-level version of this model as it binds together most of the framework's building blocks. The models reported below were created and visualized using \texttt{pyhgf} v0.2.0. Bayesian inference was performed using PyMC v5.16.2 \parencite{pymc2023}. The posterior densities and traces were plotted using Arviz v0.19.0 \parencite{Kumar2019}. The Jupyter notebook used to produce models and figures can be retrieved at \url{https://github.com/ComputationalPsychiatry/pyhgf/blob/paper/docs/paper.ipynb}.

\subsection{Generative model, forward fitting and parameter inference}

\begin{figure*}[ht]
\centering
\includegraphics[width=\textwidth]{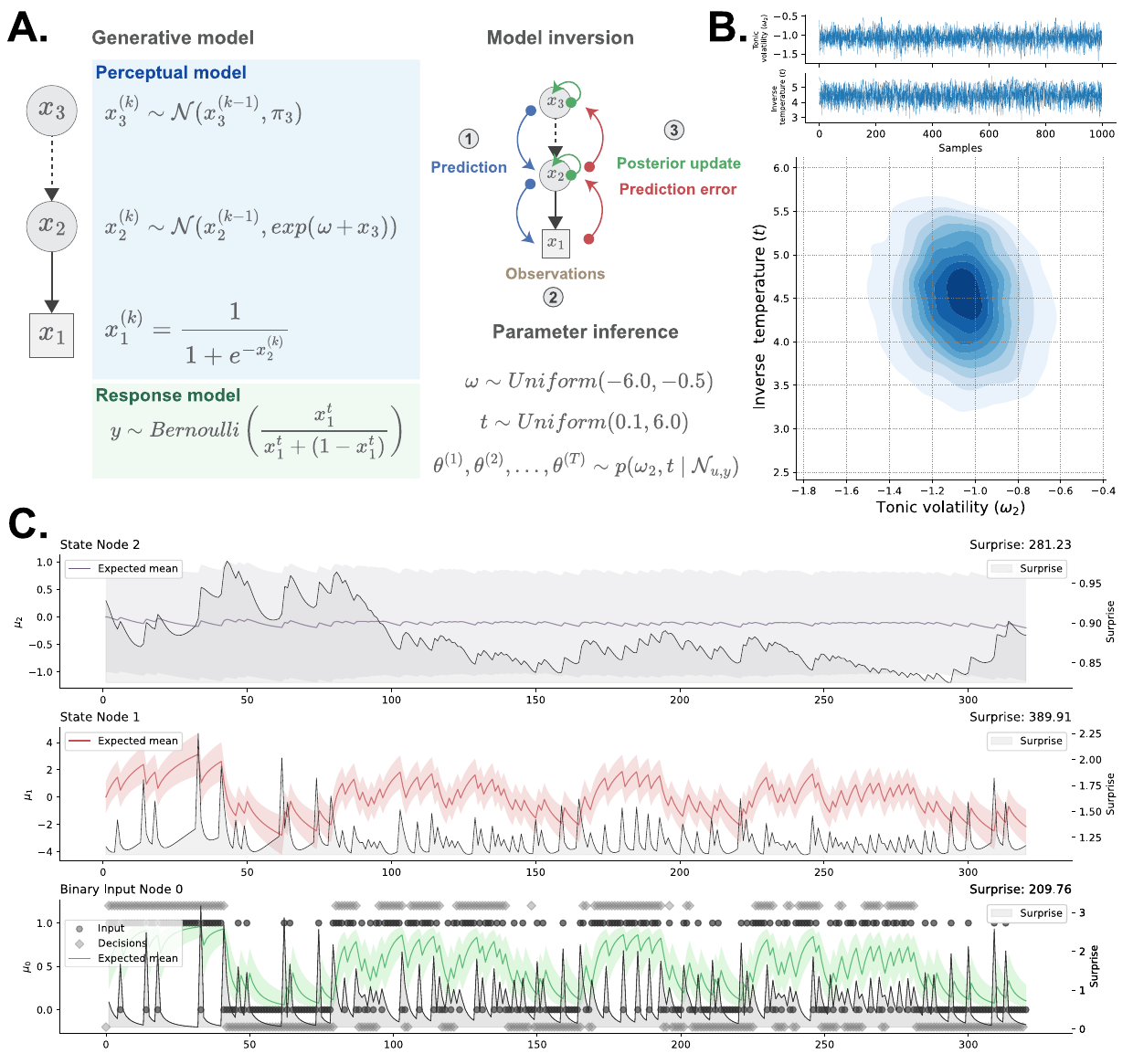}
\captionsetup{font=footnotesize}
\caption{
\color{Gray}
\textbf{Optimization and inference on the three-level Hierarchical Gaussian Filters for binary inputs. A.} 1) Graphical representation of the generative model. Square nodes represent binary state nodes and circle nodes continuous state nodes. Dashed lines indicate volatility coupling while solid lines indicate value coupling. The response model assumes a logistic sigmoid response function with an inverse temperature parameter. 2) Model fitting relies on an inversion of the generative model comprising the top-down propagation of predictions and the bottom-up propagation of prediction errors driving posterior updates. 3) Parameter inference and optimization imply a second inversion, namely that of the response model, which relies on automatic differentiation internally. The response function defines the log-probability (or negative surprise) of the observed data under the generative model. \textbf{B.} Posterior distribution of inferred parameters. Here, we inferred the value of tonic volatility ($\omega$) at the second level ($x_2$), and the inverse temperature of the response function ($t$). The upper panel displays the resulting traces (4 chains with 1000 samples), and the bottom panel is a bivariate representation of the corresponding posterior density. All outputs are compatible with PyMC \cite{pymc2023} and Arviz \cite{Kumar2019} for visualization and diagnostics. \textbf{C.} The belief trajectories across time for a model using the best parameter set from the previous steps. The grey-filled curves represent the surprise. The expected mean and precision at each level are depicted using the coloured lines and shaded areas (respectively). The plot was generated using \texttt{pyhgf}'s plotting module.}
\label{fig:inference}
\end{figure*}

The standard workflow usually starts with the creation of a Bayesian network similar to that depicted in \ref{fig:inference} (panel \textbf{A.}). This network is intended to represent the generative model of the environment as inferred by the agent. When using \texttt{pyhgf}'s built-in plotting function, the nodes that can receive observations are displayed in the lower part of the graph (the root; note that we use the terms root and leaf with reference to the \emph{inference} model where all arrows have the inverse direction to that in the \emph{generative} model depicted in panel \textbf{A.}). Here, the inputs are inserted in a binary (0 or 1) state node ($x_1$) whose value upon inversion ($\mu_1$) indicates the probability of observing the category $1$. This node's value is itself predicted by a continuous state node ($x_2$), which encodes this probability as a continuous unbounded variable after transformation by the logit function. Because the variability of this variable is assumed to change over time, the fluctuation is controlled by the third node ($x_3$) through volatility coupling, represented here using dashed lines. If the volatility is assumed to be fixed, the third node can be removed, which yields a two-level binary HGF. This graphical depiction can be generated using the internal \texttt{pyhgf} plotting library, which is convenient to visualise and debug complex networks.

\FloatBarrier

The generative model can be read as a regular Bayesian network and used to simulate a time series of belief dynamics. However, we are generally more interested in fitting this structure to existing observations and updating each node's value accordingly. This step corresponds to the first inversion of the response model and simulates the agent updating its beliefs when facing contradicting evidence. This unfolds in three parts: 1). an inference step in which a cascade of predictions is triggered from the leaves to the root of the network, 2). a new observation is received, and 3). an update step is applied through the propagation of prediction error, alternating with posterior updates, from the roots to the leaves. The resulting trajectories are influenced by the values of key node attributes, such as the tonic volatility at the second level (hereafter denoted $\omega$), of which the precision of the implied normal distribution is a monotonically decreasing function, and can be interpreted as a learning rate in this context.

The procedure above describes the perceptual model and explains how beliefs evolve in the network as new observations are made. We then assume that an agent uses available beliefs at time $k$ to inform decisions and actions. How to convert beliefs into actions depends on the problem we try to solve. Here, we assumed that the participant uses the inferred probability $\mu_1$ of  $x_1 = 1$ (moderated by logistic sigmoid noise parametrized by an inverse temperature parameter) to select its action (i.e., a prediction of 0 or 1 for the outcome). This function can generate actions from the belief trajectories, but can also quantify their likelihood if they have been recorded during the task, which can be used as a log-probability function to measure the quality of model fitting. If we assume that some attributes are not fixed, such as the tonic volatility at the second level ($\omega$) and the inverse temperature parameter ($t$), the probability density of these parameters given the observations, decision and network assumed ($P(\omega_2, t | \mathcal{N}_{u, y})$) can be recovered through MCMC sampling (see panel \textbf{B.}). By taking the average of the resulting samples, we can recover an approximation of the expected value of each variable. Here, we used four chains, each containing 1000 samples. We can then use these estimates, fit the model to the same data, and recover the belief trajectories that are most likely to explain the observed behaviours (see panel \textbf{C.}). Because the model inversion relies on a closed-form solution of the variational update, model fitting is fully deterministic, meaning that we will recover the same belief trajectories given identical inputs and network parameters.

\subsection{Bayesian multilevel modelling, parameter recovery and model comparison}

\begin{figure}[ht]
\centering
\includegraphics[width=\textwidth]{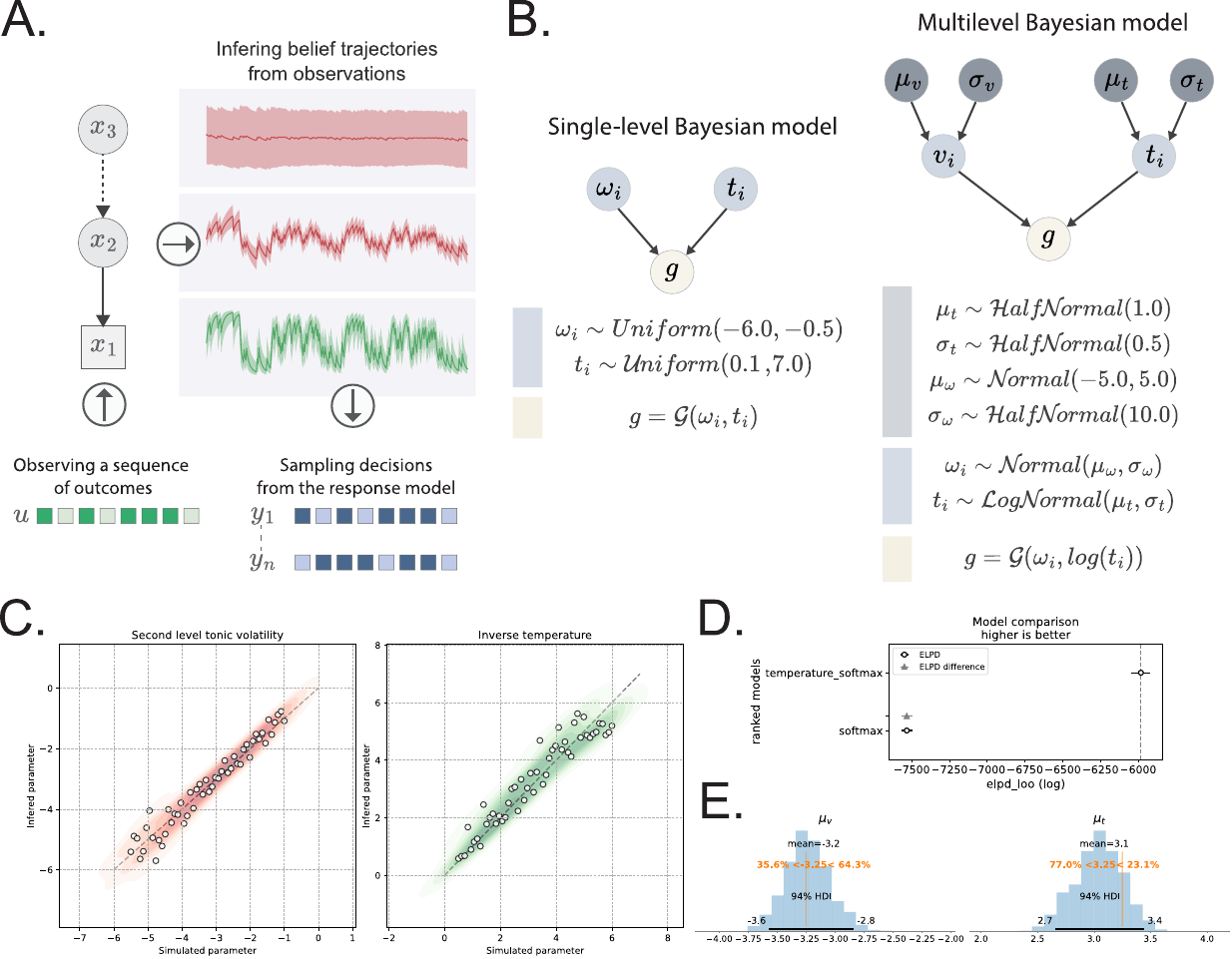}
\captionsetup{font=footnotesize}
\caption{
\color{Gray}
\textbf{Recovering computational parameters. A.} Data simulation. We used a set of observations $u$ from \cite{Iglesias2021} as environmental outcomes to simulate belief trajectories under varying values for the second level volatility ($\omega$) and the inverse temperature ($t$). The expected probability at the first level was then used to sample a vector of decisions using the same response function as described in \ref{fig:inference}.\textbf{B.} Bayesian modelling of the network's parameters using single-level and multilevel approaches. The single-level approach does not set group constraints on individual parameters and is in this case preferred for parameter recovery. We also used this version for the model comparison. The multilevel version puts hyperpriors on top of the individual parameters, enabling inference at the group level. \textbf{C.} Parameter recovery. In both panels, the horizontal axis represents the simulated value while the vertical axis represents the recovered/inferred value for tonic volatility at the second level (left) and inverse temperature (right). The dashed line shows the unit line for reference. \textbf{D.} Model comparison results. For illustration, we compared a model using a simple sigmoid function as a response function with another one using a sigmoid with an inversion temperature parameter. The plot represents the comparison between the two models based on their expected log pointwise predictive density (ELPD), which is the default recommended method for model comparison when using Arviz. \textbf{E.} Posterior estimates of group-level hyperparameters. Posterior density estimates of group mean for tonic volatility ($\omega$) and temperature ($t$). The orange vertical lines represent empirical group means, the intervals represent the 94\% highest density intervals (HDI). }
\label{fig:recovery}
\end{figure}

Cognitive neuroscience experiments involve multiple participants, and the statistical procedure requires inferring multiple parameters at once. To illustrate this practice we simulated in the second part the processing of a large simulated dataset (50 participants). We used the same vector of observations $u$, but this time generated 50 response vectors $y_i$ by sampling from the response function with varying values for the two parameters of interest ($\omega$ and $t$, see \ref{fig:recovery} panel \textbf{a.}). Using the same fitting procedure as \ref{fig:inference}, we applied this to all pairs of observations $u$ and actions $y_i$. Since this step is fully parallelized, it will benefit from hardware acceleration provided by multiple CPUs or GPUs. Note that because we assume no dependencies across participants, we would obtain the same results fitting all participants iteratively, or a group of participants in a single-level Bayesian model (see \ref{fig:recovery} panel \textbf{B.}). Next, we checked how the real values from parameters $\omega$ and $t$ could be recovered from the observations and actions alone, a practice known as parameter recovery. The recovered parameters, together with the real parameters used for the simulation of decisions, are displayed in panel \textbf{C.}. Overall, the location of the recovered parameters near the identity line suggests a good recovery from the input data. This could be further quantified, for example using correlation coefficients.

We can use different networks as generative models on the same dataset, for example, we can add or remove a node (e.g., going from a three-level to a two-level binary HGF), or use different response functions. When the structure of the model itself diverges, these models can be compared side by side using model comparison techniques that operate directly from the samples returned by the MCMC procedure \parencite{Vehtari2016}. To illustrate this we created a second model, identical to the previous one, in which the response function used a fixed inverse temperature of 1. Such a model is expected to perform worse at predicting the simulated participant's decisions, as it will miss the variability introduced by the temperature. Using the same single-level Bayesian fitting procedure, we can now compare them side-by-side using leave-one-out cross-validation (LOO), as it is implemented y default in Arviz. The results, displayed in panel \textbf{D.} show that the model where the inverse temperature was estimated instead of being fixed performed better. We emphasise that we did this here purely for illustrative purposes. Theory-agnostic model comparisons of this kind are normally a poor guide for model choice, not least because of the ease with which a particular outcome can be engineered by adjusting the priors. Instead, model choice should always be theory-driven and include analysis of the posterior predictive distribution as well as calibration of priors by prior predictive simulation.

Finally, experimenters may be interested in recovering an estimate of population parameters beyond the individual fits, for example, to compare healthy controls with patients. This requires a multilevel approach, in which participants' parameters are not statistically independent but drawn from a group distribution (see Multilevel Bayesian model in \ref{fig:recovery} panel \textbf{B.}). This constraint can help refine the estimates recovered and give more statistical power when comparing groups or conditions. We report in panel \textbf{D.} the inferred posterior density for the mean of $\omega$ and $t$ at the population level, together with the empirical means (orange vertical lines). Overall, the results suggest good reliability of the inferred posterior density, as both empirical means were included in the 94\% highest density interval (HDI) of the population estimates.

\FloatBarrier
\section{Future directions} \label{Availability}

Bayesian models of cognition have been around for decades, and frameworks like predictive coding are popular for their efficiency in modelling information processing in the central nervous system. The simplicity and modularity of the computational steps that support learning and optimization of these networks are opening new research avenues for designing structures without requiring gradient descent \parencite{Millidge:2022}, and can easily extend to various forms of processes, such as causal inference \parencite{Salvatori:2023} or temporal prediction patterns \parencite{Millidge2024}.

In this paper, we introduced \texttt{pyhgf}, a Python package that provides a generic framework for designing, manipulating and sampling dynamic networks for predictive coding. Unlike conventional neural architectures, dynamic networks as they can be implemented in \texttt{pyhgf} are tailored to react and reorganise when absorbing sensory inputs without relying on an external optimization algorithm. This framework is intentionally abstract and agnostic towards the mathematical formalism that implements inference and optimisation. We provide a complete implementation of generalised Bayesian filtering \parencite{Mathys2020} and the generalised Hierarchical Gaussian Filter \parencite{weber2023, 2011:mathys, Matthew:2014} as two important tools for predictive coding. 

The package lets the user create and manipulate arbitrarily sized networks defined as rooted trees on which scheduling or reactivity of simple local updates is applied to perform belief propagation. Critically, every step in the propagation is an in-place function receiving and returning the network itself. This supports complete plasticity and dynamics during the updates. We believe that this design allows not only the smooth manipulation of such networks but also a better dissociation between different areas of methodological development in predictive coding (e.g., the mathematical formalism versus the experimental application) so the user does not need complete expertise in all these domains to use the tools.

\begin{figure}[ht]
\centering
\includegraphics[width=160mm]{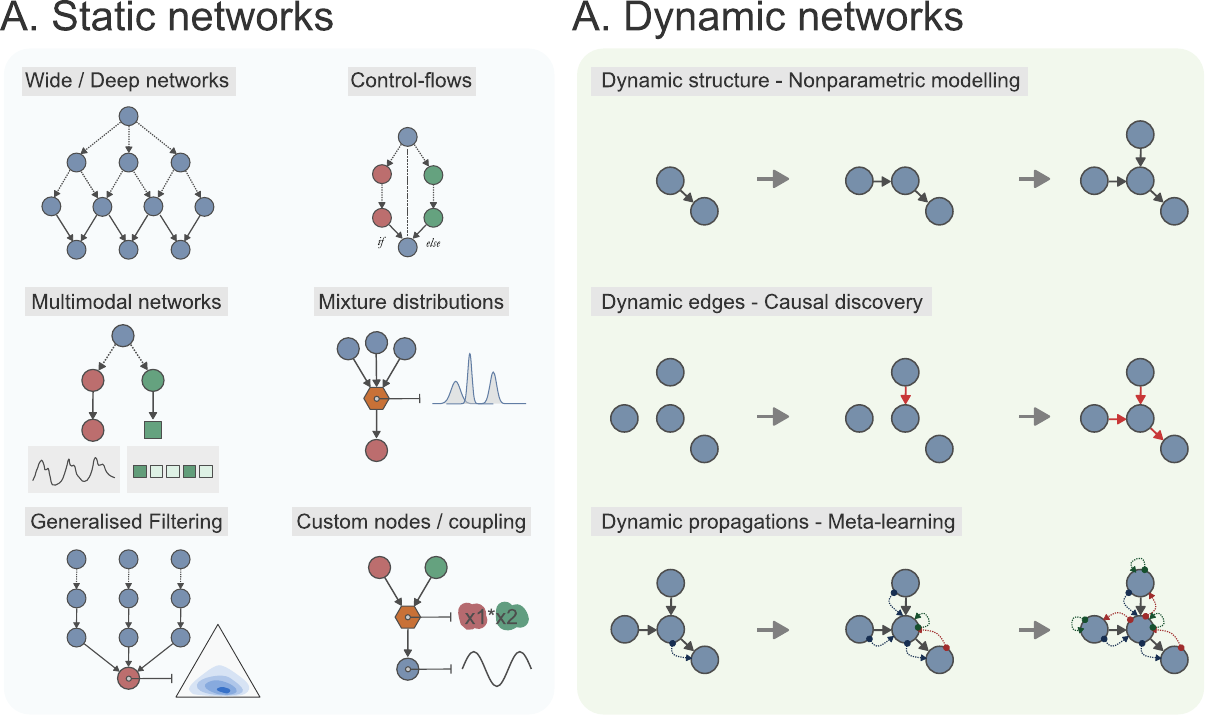}
\captionsetup{font=footnotesize}
\caption{
\color{Gray}
\textbf{Possible use cases of static and dynamic graphs.} \textbf{A.} The library supports arbitrary network structure, including deep/wide networks and multivariate dependencies/ascendencies that can handle a large number of inputs with nested hierarchical dependencies (\textit{top-left}). Belief propagation dynamics are adaptable and can implement regular control-flow statements, for example, to condition message passing on the outcome of some assertion (\textit{top-right}). Any node can observe new inputs. Branches of a network observing some inputs will specialise in their dynamics but can share volatility or value at higher levels with other branches (e.g., physiological signals and binary outcomes, \textit{middle-left}). Nodes can capture influence from multiple parents as mixture distributions for online clustering \textit{middle-right}. Any exponential family distribution can be filtered, here each node is tracing one sufficient statistic parameter, \textit{bottom-left}. The update steps can be adapted and implement custom operations either at the node level or through coupling functions \textit{bottom-right}. \textbf{B.} Dynamic graphs can update nodes' attributes and their structure, edges and propagation functions. This flexibility can be used to add an inference process that can accommodate catastrophic prediction errors and protect inference in the long term. This can imply increasing the model complexity, a principle known as Bayesian non-parametric modelling, \textit{top}. Adapting the network's edges can change the causal relationship between variables, a process known as causal discovery \textit{middle}. The propagation dynamic can expand its affordances and become more complex to improve the inference algorithm as new observations are made, a process that borrows principle from meta-learning \textit{bottom}. All the examples here depict networks assuming the context of a generalized Hierarchical Gaussian filter \parencite{weber2023} but the principles can easily be adapted to other predictive coding frameworks.}
\label{fig:new_models}
\end{figure}

We have illustrated in \ref{Results} a possible workflow when using the library, taking as an example a popular model from computational psychiatry: the three-level hierarchical Gaussian filter. In this section, we are concerned with the new possibilities offered by the framework that are not available in other libraries. These new directions have two main aspects: 1. whether the computational graphs have fixed shapes, or 2. whether the graphs are dynamic and the structure is flexible \ref{fig:new_models}. We provide examples of such implementations below.

\subsection{Generalised Bayesian filtering in static networks}

Bayesian networks are foundational models for the development of predictive coding. Importantly, the dynamic networks that we introduce here are also Bayesian networks, at least during the prediction steps. Predictive coding is more concerned with the Bayesian updating of those graphs in real time and uses prediction errors as key quantities to manipulate these updates. While static, this graphical representation is nonetheless highly relevant to modelling the nervous system \parencite{Friston2017}. One interesting aspect of this framework is that arbitrary-sized neural networks can represent arbitrary complex generative models without having to rethink an entire algorithm: simple and principled local updates are generalized across the network to approximate variational inference. While nodes are implicitly tracking one parameter value through unidimensional normal distributions, arbitrarily complex networks can combine any exponential family distribution by using multiple nodes to track sufficient statistics \parencite{Mathys2020}. This approach in itself offers considerable modularity in real-time probabilistic modelling, as the development of complex variational update algorithms can be replaced by the manipulation of nodes in a large network. We believe that the flexibility and modularity proposed by the framework will continue to extend applications of the Hierarchical Gaussian Filter on various filtering problems \parencite{Senoz2020, Senoz2021}. 

One straightforward consequence is that networks can be extended in several dimensions: not only horizontally (by adding more input nodes) and vertically (adding more parent nodes), but also through multivariate descendency (one parent node influencing many child nodes) or ascendency (multiple parent nodes influencing one child node). Extensions of the model can include for example multiple inputs design, mixture of distributions or combination of parents influence into custom operations. By setting dependencies between branches of a network that are oriented toward the processing of certain kinds of signals, it is for example possible to bring together the high dimensionality and multimodality of observation usually recorded in cognitive neuroscience. This can find direct applications in situations where neural physiological recordings should be related to behavioural outcomes and decisions. Here, a branch of the network can filter one stream of physiological signals (e.g. respiration, heart rate, EEG or fMRI signals) while another branch is used as a decision-making process (see the middle-left panel in \ref{fig:new_models} \textbf{a.}). Such situations are pervasive in the field, and statistical analyses are often carried out by correlation of independent modality-dedicated methods. whereas here, a multi-branch HGF can bind together the mutual influence of behaviours and physiology in a consistent model. 

\subsection{Structure learning, causal inference and meta-learning in dynamic networks}

Besides the large flexibility provided by \texttt{pyhgf} regarding the connectivity of networks, the malleability of structural variables during belief propagation also opens new avenues. Predictive coding networks use precision-weighted prediction errors to drive posterior updates. Updates are proportional to the error computed at a lower level and even when large errors are observed the network needs to accommodate this new evidence gracefully. If it does not, this can have disastrous consequences for inference, for example, if surprise increases unreasonably because of overly precise predictions, as can happen with improper precision weighting, or simply when receiving observations that are very far from the expected distribution. It is therefore essential for these systems to implement firewalls that could contain the propagation of catastrophic updates and preserve past learnings from complete erasure. Here, we suggest that the dynamic reconfiguration of the networks can serve as an alternative to belief updating under disproportionate surprise, and preserve the model's integrity. While this idea arises from the application of biological principles to Bayesian networks, its implementation can express a variety of other popular learning and inference algorithms, namely structure learning, causal discovery and meta-learning.

Self-assembling and self-growing neural networks are a new, active, and rapidly advancing area of research in deep reinforcement learning \parencite{Najarro2023}. These approaches stress the notion that biological as well as artificial agents should implement a flexible reorganization of their generative model to support lifelong learning \parencite{Kudithipudi2022}, a feature that also remains absent in many applications of predictive coding to date. While several methods have been developed which extend conventional deep neural networks, we propose here that dynamic networks can naturally live up to this principle by simply adding or removing nodes during belief propagation (see the top panel in \ref{fig:new_models} \textbf{B.}). This feature leverages the flexibility of a predictive coding framework to implement branching, splitting, and merging behaviours under prediction errors, which can be a way to address non-parametric modelling problems. It is worth noting here that the dynamic reshaping of networks in real-time contributes to both the memory and energy efficiency of the system, ensuring that at each time step, only an optimal number of nodes are carried over. The resulting structure then works as a comprehensive encoding of relevant past environmental volatility, building on the notion that the shape of the system, and beyond that the agent's body itself, is an active part of the inference process.

This modulation of the network structure can also concern only the edges, without adding or removing nodes. Because dynamic networks behave like Bayesian networks during the prediction steps, editing the edges equates to modulating the causal relationship between variables, which is a way to express causal discovery algorithms. In this context, the agent tries to recover the causal graph from observation only (see the middle panel in \ref{fig:new_models} \textbf{B.}). It is widely assumed that inference over causality in general plays an important part in the development of human and animal cognition \parencite{Goddu2024}, with applications ranging from improving conventional reinforcement learning methods \parencite{Deng2023} to the analytics of structured texts and narrative \parencite{Chen2024}. In the framework of \texttt{pyhgf}, causal influence can be inferred in real-time, which again is a way to optimise memory and energy consumption and open possibilities for the estimation of time-varying causal relationships, or even degrees of volatility in causal influence, a notion that could find several applications in computational psychiatry.

Finally, the structural malleability of dynamic networks can target the belief propagation dynamic itself. Belief propagation is a sequence of functions, therefore both the sequence and the functions can be adapted to improve inference when aggregating more observations. Such plasticity emerges naturally in biological networks, but it is obscured by the software implementation when it comes to artificial neural networks. In \texttt{pyhgf}, update functions themselves are part of the sufficient descriptive variables of a network passed during belief updates, meaning that inference can edit them in real time to improve predictive accuracy (see the lower panel in \ref{fig:new_models} \textbf{B.}). This approach borrows principles from both Bayesian non-parametric modelling approaches such as Gaussian Processes that update posterior distribution over a function, and from meta-learning, a generalisation of reinforcement learning where an agent tries to learn a learning algorithm \parencite{Binz2023}. The propagation itself can increase its range (i.e., propagate beliefs into deeper portions of the network), as well as the update function, which can be done in a deterministic way by setting a prior over a range of coupling functions, or by the automated generation of code, which is now made easier given the new application of Large Language Models in this domain \parencite{willard:2023}.

In addition to these core computational internal changes, and closer to active inference models, the package also supports the modular extension of the response functions used by the agent (i.e., the beliefs-to-action function). This framework is central for reinforcement learning applications and often requires custom response functions tailored to specific tasks. Examples of how to use custom response functions can be found in the online documentation.

\FloatBarrier
\section{Conclusion}

In this paper, we have introduced \texttt{pyhgf}, a neural network library for predictive coding with a focus on generalised Bayesian filtering and the generalised hierarchical Gaussian filter (HGF). We described how the modular definition of neural networks supporting the scheduling of update steps can serve as a generic framework for models relying on the propagation of simple local computations through a hierarchy of layers, such as in predictive coding neural networks. One part of the API is dedicated to the flexible development of dynamic networks, while the second part is oriented towards high-level use and parameter inference, as typically requested for computational neuroscience studies. Together, we hope that this toolbox will help and strengthen the application of predictive neural networks in computational psychiatry, and open new designs in artificial intelligence towards hybrid and complex models of cognition that build on the principled computations derived from predictive coding. \texttt{pyhgf} can be installed from the Python Package Index (\url{https://pypi.org/project/pyhgf/}) and the source code is hosted on GitHub under the following public repository: \url{https://github.com/ComputationalPsychiatry/pyhgf}. The documentation for the most recent version is accessible at the following link: \url {https://ComputationalPsychiatry.github.io/pyhgf/index.html}. The documentation hosts extensive tutorials, examples, and use cases with applications on signal processing, reinforcement learning, and computational psychiatry. We point interested readers to these resources for a deeper practical introduction to the library.

\printglossary

\clearpage
\pagestyle{empty}
\newgeometry{bottom=.5in,top=.5in,left=.5in,right=.5in}  
\twocolumn
\renewcommand*{\bibfont}{\small}  
\printbibliography

@article{2011:mathys,
author = {Mathys, Christoph},
doi = {10.3389/fnhum.2011.00039},
issn = {16625161},
journal = {Frontiers in Human Neuroscience},
number = {May},
pages = {1--20},
pmid = {21629826},
title = {{A Bayesian foundation for individual learning under uncertainty}},
url = {http://journal.frontiersin.org/article/10.3389/fnhum.2011.00039/abstract},
volume = {5},
year = {2011}
}

@ARTICLE{2014:mathys,
AUTHOR={Mathys, Christoph and Lomakina, Ekaterina I. and Daunizeau, Jean and Iglesias, Sandra and Brodersen, Kay H. and Friston, Karl J. and Stephan, Klaas E.},
TITLE={Uncertainty in perception and the Hierarchical Gaussian Filter},
JOURNAL={Frontiers in Human Neuroscience},
VOLUME={8},
YEAR={2014},
URL={https://www.frontiersin.org/articles/10.3389/fnhum.2014.00825},
DOI={10.3389/fnhum.2014.00825},
ISSN={1662-5161}
}

@article{Iglesias2021,
  doi = {10.1016/j.neuroimage.2020.117590},
  url = {https://doi.org/10.1016/j.neuroimage.2020.117590},
  year = {2021},
  month = feb,
  publisher = {Elsevier {BV}},
  volume = {226},
  pages = {117590},
  author = {Sandra Iglesias and Lars Kasper and Samuel J. Harrison and Robert Manka and Christoph Mathys and Klaas E. Stephan},
  title = {Cholinergic and dopaminergic effects on prediction error and uncertainty responses during sensory associative learning},
  journal = {{NeuroImage}}
}

@inproceedings{Senoz2020,
  doi = {10.1109/isit44484.2020.9173980},
  url = {https://doi.org/10.1109/isit44484.2020.9173980},
  year = {2020},
  month = jun,
  publisher = {{IEEE}},
  author = {Ismail Senoz and Bert de Vries},
  title = {Online Message Passing-based Inference in the Hierarchical Gaussian Filter},
  booktitle = {2020 {IEEE} International Symposium on Information Theory ({ISIT})}
}

@inproceedings{Senoz2021,
  doi = {10.1109/isit45174.2021.9518229},
  url = {https://doi.org/10.1109/isit45174.2021.9518229},
  year = {2021},
  month = jul,
  publisher = {{IEEE}},
  author = {Ismail Senoz and Albert Podusenko and Semih Akbayrak and Christoph Mathys and Bert de Vries},
  title = {The Switching Hierarchical Gaussian Filter},
  booktitle = {2021 {IEEE} International Symposium on Information Theory ({ISIT})}
}

@article{Powers2017,
  doi = {10.1126/science.aan3458},
  url = {https://doi.org/10.1126/science.aan3458},
  year = {2017},
  month = aug,
  publisher = {American Association for the Advancement of Science ({AAAS})},
  volume = {357},
  number = {6351},
  pages = {596--600},
  author = {A. R. Powers and Chistoph Mathys and P. R. Corlett},
  title = {Pavlovian conditioning{\textendash}induced hallucinations result from overweighting of perceptual priors},
  journal = {Science}
}

@article{Lawson2017,
  doi = {10.1038/nn.4615},
  url = {https://doi.org/10.1038/nn.4615},
  year = {2017},
  month = jul,
  publisher = {Springer Science and Business Media {LLC}},
  volume = {20},
  number = {9},
  pages = {1293--1299},
  author = {Rebecca P Lawson and Christoph Mathys and Geraint Rees},
  title = {Adults with autism overestimate the volatility of the sensory environment},
  journal = {Nature Neuroscience}
}

@incollection{Mathys2020,
  doi = {10.1007/978-3-030-64919-7_7},
  url = {https://doi.org/10.1007/978-3-030-64919-7_7},
  year = {2020},
  publisher = {Springer International Publishing},
  pages = {52--58},
  author = {Christoph Mathys and Lilian Weber},
  title = {Hierarchical Gaussian Filtering of Sufficient Statistic Time Series for Active Inference},
  booktitle = {Active Inference}
}

@article{Frssle2021,
  doi = {10.3389/fpsyt.2021.680811},
  url = {https://doi.org/10.3389/fpsyt.2021.680811},
  year = {2021},
  month = jun,
  publisher = {Frontiers Media {SA}},
  volume = {12},
  author = {Stefan Fr\"{a}ssle and Eduardo A. Aponte and Saskia Bollmann and Kay H. Brodersen and Cao T. Do and Olivia K. Harrison and Samuel J. Harrison and Jakob Heinzle and Sandra Iglesias and Lars Kasper and Ekaterina I. Lomakina and Christoph Mathys and Matthias M\"{u}ller-Schrader and In{\^{e}}s Pereira and Frederike H. Petzschner and Sudhir Raman and Dario Sch\"{o}bi and Birte Toussaint and Lilian A. Weber and Yu Yao and Klaas E. Stephan},
  title = {{TAPAS}: An Open-Source Software Package for Translational Neuromodeling and Computational Psychiatry},
  journal = {Frontiers in Psychiatry}
}

@misc{weber2023,
  doi = {10.48550/ARXIV.2305.10937},
  url = {https://arxiv.org/abs/2305.10937},
  author = {Weber,  Lilian Aline and Waade,  Peter Thestrup and Legrand,  Nicolas and Møller,  Anna Hedvig and Stephan,  Klaas Enno and Mathys,  Christoph},
  title = {The generalized Hierarchical Gaussian Filter},
  publisher = {arXiv},
  year = {2023},
  copyright = {Creative Commons Attribution Non Commercial No Derivatives 4.0 International}
}

@software{deepmind2020jax,
  title = {The {D}eep{M}ind {JAX} {E}cosystem},
  author = {Babuschkin, Igor and Baumli, Kate and Bell, Alison and Bhupatiraju, Surya and Bruce, Jake and Buchlovsky, Peter and Budden, David and Cai, Trevor and Clark, Aidan and Danihelka, Ivo and Dedieu, Antoine and Fantacci, Claudio and Godwin, Jonathan and Jones, Chris and Hemsley, Ross and Hennigan, Tom and Hessel, Matteo and Hou, Shaobo and Kapturowski, Steven and Keck, Thomas and Kemaev, Iurii and King, Michael and Kunesch, Markus and Martens, Lena and Merzic, Hamza and Mikulik, Vladimir and Norman, Tamara and Papamakarios, George and Quan, John and Ring, Roman and Ruiz, Francisco and Sanchez, Alvaro and Sartran, Laurent and Schneider, Rosalia and Sezener, Eren and Spencer, Stephen and Srinivasan, Srivatsan and Stanojevi\'{c}, Milo\v{s} and Stokowiec, Wojciech and Wang, Luyu and Zhou, Guangyao and Viola, Fabio},
  url = {http://github.com/deepmind},
  year = {2020},
}

@misc{hoffman2020,
  doi = {10.48550/ARXIV.2006.00979},
  url = {https://arxiv.org/abs/2006.00979},
  author = {Hoffman,  Matthew W. and Shahriari,  Bobak and Aslanides,  John and Barth-Maron,  Gabriel and Momchev,  Nikola and Sinopalnikov,  Danila and Stańczyk,  Piotr and Ramos,  Sabela and Raichuk,  Anton and Vincent,  Damien and Hussenot,  Léonard and Dadashi,  Robert and Dulac-Arnold,  Gabriel and Orsini,  Manu and Jacq,  Alexis and Ferret,  Johan and Vieillard,  Nino and Ghasemipour,  Seyed Kamyar Seyed and Girgin,  Sertan and Pietquin,  Olivier and Behbahani,  Feryal and Norman,  Tamara and Abdolmaleki,  Abbas and Cassirer,  Albin and Yang,  Fan and Baumli,  Kate and Henderson,  Sarah and Friesen,  Abe and Haroun,  Ruba and Novikov,  Alex and Colmenarejo,  Sergio Gómez and Cabi,  Serkan and Gulcehre,  Caglar and Paine,  Tom Le and Srinivasan,  Srivatsan and Cowie,  Andrew and Wang,  Ziyu and Piot,  Bilal and de Freitas,  Nando},
  title = {Acme: A Research Framework for Distributed Reinforcement Learning},
  publisher = {arXiv},
  year = {2020},
  copyright = {arXiv.org perpetual,  non-exclusive license}
}

@software{jraph2020,
  author = {Jonathan Godwin and Thomas Keck and Peter Battaglia and Victor Bapst and Thomas Kipf and Yujia Li and Kimberly Stachenfeld and Petar Veli\v{c}kovi\'{c} and Alvaro Sanchez-Gonzalez},
  title = {{J}raph: {A} library for graph neural networks in jax.},
  url = {http://github.com/deepmind/jraph},
  version = {0.0.1.dev},
  year = {2020},
}

@software{jax2018github,
  author = {James Bradbury and Roy Frostig and Peter Hawkins and Matthew James Johnson and Chris Leary and Dougal Maclaurin and George Necula and Adam Paszke and Jake Vander{P}las and Skye Wanderman-{M}ilne and Qiao Zhang},
  title = {{JAX}: composable transformations of {P}ython+{N}um{P}y programs},
  url = {http://github.com/google/jax},
  version = {0.3.13},
  year = {2018},
}

@misc{Kidger2021,
  doi = {10.48550/ARXIV.2111.00254},
  url = {https://arxiv.org/abs/2111.00254},
  author = {Kidger,  Patrick and Garcia,  Cristian},
  title = {Equinox: neural networks in JAX via callable PyTrees and filtered transformations},
  publisher = {arXiv},
  year = {2021},
  copyright = {Creative Commons Attribution 4.0 International}
}

@article{pymc2023,
  title={PyMC: A Modern and Comprehensive Probabilistic Programming Framework in Python},
  author={Abril-Pla Oriol and Andreani Virgile and Carroll Colin and Dong Larry and Fonnesbeck Christopher J. and Kochurov Maxim and Kumar Ravin and Lao Jupeng and Luhmann Christian C. and Martin Osvaldo A. and Osthege Michael and Vieira Ricardo and Wiecki Thomas and Zinkov Robert},
  journal = {{PeerJ} Computer Science},
  publisher = {{PeerJ}},
  volume={9},
  pages={e1516},
  year={2023},
  doi={10.7717/peerj-cs.1516}
}

@article{Kumar2019,
  doi = {10.21105/joss.01143},
  url = {https://doi.org/10.21105/joss.01143},
  year = {2019},
  month = jan,
  publisher = {The Open Journal},
  volume = {4},
  number = {33},
  pages = {1143},
  author = {Ravin Kumar and Colin Carroll and Ari Hartikainen and Osvaldo Martin},
  title = {{ArviZ} a unified library for exploratory analysis of Bayesian models in Python},
  journal = {Journal of Open Source Software}
}

@article{Friston2017,
  doi = {10.1162/netn_a_00018},
  url = {https://doi.org/10.1162/netn_a_00018},
  year = {2017},
  month = dec,
  publisher = {{MIT} Press},
  volume = {1},
  number = {4},
  pages = {381--414},
  author = {Karl J. Friston and Thomas Parr and Bert de Vries},
  title = {The graphical brain: Belief propagation and active inference},
  journal = {Network Neuroscience}
}

@article{Corlett2019,
  doi = {10.1016/j.tics.2018.12.001},
  url = {https://doi.org/10.1016/j.tics.2018.12.001},
  year = {2019},
  month = feb,
  publisher = {Elsevier {BV}},
  volume = {23},
  number = {2},
  pages = {114--127},
  author = {Philip R. Corlett and Guillermo Horga and Paul C. Fletcher and Ben Alderson-Day and Katharina Schmack and Albert R. Powers},
  title = {Hallucinations and Strong Priors},
  journal = {Trends in Cognitive Sciences}
}

@article{Reed2020,
  doi = {10.7554/elife.56345},
  url = {https://doi.org/10.7554/elife.56345},
  year = {2020},
  month = may,
  publisher = {{eLife} Sciences Publications,  Ltd},
  volume = {9},
  author = {Erin J Reed and Stefan Uddenberg and Praveen Suthaharan and Christoph D Mathys and Jane R Taylor and Stephanie Mary Groman and Philip R Corlett},
  title = {Paranoia as a deficit in non-social belief updating},
  journal = {{eLife}}
}

@misc{Najarro2023,
  doi = {10.48550/ARXIV.2307.08197},
  url = {https://arxiv.org/abs/2307.08197},
  author = {Najarro,  Elias and Sudhakaran,  Shyam and Risi,  Sebastian},
  title = {Towards Self-Assembling Artificial Neural Networks through Neural Developmental Programs},
  publisher = {arXiv},
  year = {2023},
  copyright = {Creative Commons Attribution 4.0 International}
}

@article{Kudithipudi2022,
  doi = {10.1038/s42256-022-00452-0},
  url = {https://doi.org/10.1038/s42256-022-00452-0},
  year = {2022},
  month = mar,
  publisher = {Springer Science and Business Media {LLC}},
  volume = {4},
  number = {3},
  pages = {196--210},
  author = {Dhireesha Kudithipudi and Mario Aguilar-Simon and Jonathan Babb and Maxim Bazhenov and Douglas Blackiston and Josh Bongard and Andrew P. Brna and Suraj Chakravarthi Raja and Nick Cheney and Jeff Clune and Anurag Daram and Stefano Fusi and Peter Helfer and Leslie Kay and Nicholas Ketz and Zsolt Kira and Soheil Kolouri and Jeffrey L. Krichmar and Sam Kriegman and Michael Levin and Sandeep Madireddy and Santosh Manicka and Ali Marjaninejad and Bruce McNaughton and Risto Miikkulainen and Zaneta Navratilova and Tej Pandit and Alice Parker and Praveen K. Pilly and Sebastian Risi and Terrence J. Sejnowski and Andrea Soltoggio and Nicholas Soures and Andreas S. Tolias and Dar{\'{\i}}o Urbina-Mel{\'{e}}ndez and Francisco J. Valero-Cuevas and Gido M. van de Ven and Joshua T. Vogelstein and Felix Wang and Ron Weiss and Angel Yanguas-Gil and Xinyun Zou and Hava Siegelmann},
  title = {Biological underpinnings for lifelong learning machines},
  journal = {Nature Machine Intelligence}
}

@BOOK{Ji:2023,
  title     = "Bayesian models of perception and action",
  author    = "Ji, Wei and Kording, Konrad Paul",
  publisher = "MIT Press",
  month     =  aug,
  year      =  2023,
  address   = "London, England",
  language  = "en"
}

@ARTICLE{Rao:1999,
  title     = "Predictive coding in the visual cortex: a functional
               interpretation of some extra-classical receptive-field effects",
  author    = "Rao, R P and Ballard, D H",
  journal   = "Nat. Neurosci.",
  publisher = "Springer Science and Business Media LLC",
  volume    =  2,
  number    =  1,
  pages     = "79--87",
  month     =  jan,
  year      =  1999,
  language  = "en"
}

@ARTICLE{Friston:2005,
  title     = "A theory of cortical responses",
  author    = "Friston, Karl J.",
  journal   = "Philos. Trans. R. Soc. Lond. B Biol. Sci.",
  publisher = "The Royal Society",
  volume    =  360,
  number    =  1456,
  pages     = "815--836",
  month     =  apr,
  year      =  2005,
  language  = "en"
}

@misc{millidge:predictivecodingreview,
      title={Predictive Coding: a Theoretical and Experimental Review}, 
      author={Beren Millidge and Anil Seth and Christopher L Buckley},
      year={2022},
      eprint={2107.12979},
      archivePrefix={arXiv},
      primaryClass={cs.AI},
      url={https://arxiv.org/abs/2107.12979}, 
}

@ARTICLE{Millidge:2022,
  title         = "Predictive coding: Towards a future of deep learning beyond
                   backpropagation?",
  author        = "Millidge, Beren and Salvatori, Tommaso and Song, Yuhang and
                   Bogacz, Rafal and Lukasiewicz, Thomas",
  month         =  feb,
  year          =  2022,
  copyright     = "http://arxiv.org/licenses/nonexclusive-distrib/1.0/",
  archivePrefix = "arXiv",
  primaryClass  = "cs.NE",
  eprint        = "2202.09467"
}

@article{Huys2016,
  title = {Computational psychiatry as a bridge from neuroscience to clinical applications},
  volume = {19},
  ISSN = {1546-1726},
  url = {http://dx.doi.org/10.1038/nn.4238},
  DOI = {10.1038/nn.4238},
  number = {3},
  journal = {Nature Neuroscience},
  publisher = {Springer Science and Business Media LLC},
  author = {Huys,  Quentin J M and Maia,  Tiago V and Frank,  Michael J},
  year = {2016},
  month = feb,
  pages = {404–413}
}

@article{Friston2008,
  title = {Hierarchical Models in the Brain},
  volume = {4},
  ISSN = {1553-7358},
  url = {http://dx.doi.org/10.1371/journal.pcbi.1000211},
  DOI = {10.1371/journal.pcbi.1000211},
  number = {11},
  journal = {PLoS Computational Biology},
  publisher = {Public Library of Science (PLoS)},
  author = {Friston,  Karl J.},
  editor = {Sporns,  Olaf},
  year = {2008},
  month = nov,
  pages = {e1000211}
}

@article{Ororbia2022,
  title = {The neural coding framework for learning generative models},
  volume = {13},
  ISSN = {2041-1723},
  url = {http://dx.doi.org/10.1038/s41467-022-29632-7},
  DOI = {10.1038/s41467-022-29632-7},
  number = {1},
  journal = {Nature Communications},
  publisher = {Springer Science and Business Media LLC},
  author = {Ororbia,  Alexander and Kifer,  Daniel},
  year = {2022},
  month = apr 
}

@article{Sandhu2023,
title = {Transdiagnostic computations of uncertainty: towards a new lens on intolerance of uncertainty},
journal = {Neuroscience \& Biobehavioral Reviews},
volume = {148},
pages = {105123},
year = {2023},
issn = {0149-7634},
doi = {https://doi.org/10.1016/j.neubiorev.2023.105123},
url = {https://www.sciencedirect.com/science/article/pii/S0149763423000921},
author = {Timothy R. Sandhu and Bowen Xiao and Rebecca P. Lawson}
}

@article{Rumelhart1986,
  title = {Learning representations by back-propagating errors},
  volume = {323},
  ISSN = {1476-4687},
  url = {http://dx.doi.org/10.1038/323533a0},
  DOI = {10.1038/323533a0},
  number = {6088},
  journal = {Nature},
  publisher = {Springer Science and Business Media LLC},
  author = {Rumelhart,  David E. and Hinton,  Geoffrey E. and Williams,  Ronald J.},
  year = {1986},
  month = oct,
  pages = {533–536}
}

@article{DaCosta2022,
  title = {How Active Inference Could Help Revolutionise Robotics},
  volume = {24},
  ISSN = {1099-4300},
  url = {http://dx.doi.org/10.3390/e24030361},
  DOI = {10.3390/e24030361},
  number = {3},
  journal = {Entropy},
  publisher = {MDPI AG},
  author = {Da Costa,  Lancelot and Lanillos,  Pablo and Sajid,  Noor and Friston,  Karl J. and Khan,  Shujhat},
  year = {2022},
  month = mar,
  pages = {361}
}

@ARTICLE{Betancourt2017,
  title     = "A conceptual introduction to Hamiltonian Monte Carlo",
  author    = "Betancourt, Michael",
  publisher = "arXiv",
  year      =  2017
}

@misc{Salvatori:2023,
  doi = {10.48550/ARXIV.2306.15479},
  url = {https://arxiv.org/abs/2306.15479},
  author = {Salvatori,  Tommaso and Pinchetti,  Luca and M'Charrak,  Amine and Millidge,  Beren and Lukasiewicz,  Thomas},
  title = {Causal Inference via Predictive Coding},
  publisher = {arXiv},
  year = {2023},
  copyright = {arXiv.org perpetual,  non-exclusive license}
}

@article{DeDomenico2023,
  title = {More is different in real-world multilayer networks},
  volume = {19},
  ISSN = {1745-2481},
  url = {http://dx.doi.org/10.1038/s41567-023-02132-1},
  DOI = {10.1038/s41567-023-02132-1},
  number = {9},
  journal = {Nature Physics},
  publisher = {Springer Science and Business Media LLC},
  author = {De Domenico,  Manlio},
  year = {2023},
  month = aug,
  pages = {1247–1262}
}

@article{Matthew:2014,
author = {Homan, Matthew D. and Gelman, Andrew},
title = {The No-U-turn sampler: adaptively setting path lengths in Hamiltonian Monte Carlo},
year = {2014},
issue_date = {January 2014},
publisher = {JMLR.org},
volume = {15},
number = {1},
issn = {1532-4435},
journal = {J. Mach. Learn. Res.},
month = jan,
pages = {1593–1623},
numpages = {31}
}

@software{flax2020github,
  author = {Jonathan Heek and Anselm Levskaya and Avital Oliver and Marvin Ritter and Bertrand Rondepierre and Andreas Steiner and Marc van {Z}ee},
  title = {{F}lax: A neural network library and ecosystem for {JAX}},
  url = {http://github.com/google/flax},
  version = {0.7.5},
  year = {2023},
}

@article{Friston2022,
  title = {Computational psychiatry: from synapses to sentience},
  volume = {28},
  ISSN = {1476-5578},
  url = {http://dx.doi.org/10.1038/s41380-022-01743-z},
  DOI = {10.1038/s41380-022-01743-z},
  number = {1},
  journal = {Molecular Psychiatry},
  publisher = {Springer Science and Business Media LLC},
  author = {Friston,  Karl J.},
  year = {2022},
  month = sep,
  pages = {256–268}
}

@article{Mikulasch2023,
  title = {Where is the error? Hierarchical predictive coding through dendritic error computation},
  volume = {46},
  ISSN = {0166-2236},
  url = {http://dx.doi.org/10.1016/j.tins.2022.09.007},
  DOI = {10.1016/j.tins.2022.09.007},
  number = {1},
  journal = {Trends in Neurosciences},
  publisher = {Elsevier BV},
  author = {Mikulasch,  Fabian A. and Rudelt,  Lucas and Wibral,  Michael and Priesemann,  Viola},
  year = {2023},
  month = jan,
  pages = {45–59}
}

@misc{tensorflow2015,
title={ {TensorFlow}: Large-Scale Machine Learning on Heterogeneous Systems},
url={https://www.tensorflow.org/},
note={Software available from tensorflow.org},
author={
    Mart\'{i}n~Abadi and
    Ashish~Agarwal and
    Paul~Barham and
    Eugene~Brevdo and
    Zhifeng~Chen and
    Craig~Citro and
    Greg~S.~Corrado and
    Andy~Davis and
    Jeffrey~Dean and
    Matthieu~Devin and
    Sanjay~Ghemawat and
    Ian~Goodfellow and
    Andrew~Harp and
    Geoffrey~Irving and
    Michael~Isard and
    Yangqing Jia and
    Rafal~Jozefowicz and
    Lukasz~Kaiser and
    Manjunath~Kudlur and
    Josh~Levenberg and
    Dandelion~Man\'{e} and
    Rajat~Monga and
    Sherry~Moore and
    Derek~Murray and
    Chris~Olah and
    Mike~Schuster and
    Jonathon~Shlens and
    Benoit~Steiner and
    Ilya~Sutskever and
    Kunal~Talwar and
    Paul~Tucker and
    Vincent~Vanhoucke and
    Vijay~Vasudevan and
    Fernanda~Vi\'{e}gas and
    Oriol~Vinyals and
    Pete~Warden and
    Martin~Wattenberg and
    Martin~Wicke and
    Yuan~Yu and
    Xiaoqiang~Zheng},
  year={2015},
}

@misc{pytorch,
  doi = {10.48550/ARXIV.1912.01703},
  url = {https://arxiv.org/abs/1912.01703},
  author = {Paszke,  Adam and Gross,  Sam and Massa,  Francisco and Lerer,  Adam and Bradbury,  James and Chanan,  Gregory and Killeen,  Trevor and Lin,  Zeming and Gimelshein,  Natalia and Antiga,  Luca and Desmaison,  Alban and K\"{o}pf,  Andreas and Yang,  Edward and DeVito,  Zach and Raison,  Martin and Tejani,  Alykhan and Chilamkurthy,  Sasank and Steiner,  Benoit and Fang,  Lu and Bai,  Junjie and Chintala,  Soumith},
  title = {PyTorch: An Imperative Style,  High-Performance Deep Learning Library},
  publisher = {arXiv},
  year = {2019},
  copyright = {arXiv.org perpetual,  non-exclusive license}
}

@article{Oliver1952,
  title = {Efficient Coding},
  volume = {31},
  ISSN = {0005-8580},
  url = {http://dx.doi.org/10.1002/j.1538-7305.1952.tb01403.x},
  DOI = {10.1002/j.1538-7305.1952.tb01403.x},
  number = {4},
  journal = {Bell System Technical Journal},
  publisher = {Institute of Electrical and Electronics Engineers (IEEE)},
  author = {Oliver,  B. M.},
  year = {1952},
  month = jul,
  pages = {724–750}
}

@article{Daunizeau2010,
  title = {Observing the Observer (I): Meta-Bayesian Models of Learning and Decision-Making},
  volume = {5},
  ISSN = {1932-6203},
  url = {http://dx.doi.org/10.1371/journal.pone.0015554},
  DOI = {10.1371/journal.pone.0015554},
  number = {12},
  journal = {PLoS ONE},
  publisher = {Public Library of Science (PLoS)},
  author = {Daunizeau,  Jean and den Ouden,  Hanneke E. M. and Pessiglione,  Matthias and Kiebel,  Stefan J. and Stephan,  Klaas E. and Friston,  Karl J.},
  editor = {Sporns,  Olaf},
  year = {2010},
  month = dec,
  pages = {e15554}
}

@misc{PyTorchGeometric,
  doi = {10.48550/ARXIV.1903.02428},
  url = {https://arxiv.org/abs/1903.02428},
  author = {Fey,  Matthias and Lenssen,  Jan Eric},
  title = {Fast Graph Representation Learning with PyTorch Geometric},
  publisher = {arXiv},
  year = {2019},
  copyright = {arXiv.org perpetual,  non-exclusive license}
}

@article{rust,
author = {Matsakis, Nicholas D. and Klock, Felix S.},
title = {The rust language},
year = {2014},
issue_date = {December 2014},
publisher = {Association for Computing Machinery},
address = {New York, NY, USA},
volume = {34},
number = {3},
issn = {1094-3641},
url = {https://doi.org/10.1145/2692956.2663188},
doi = {10.1145/2692956.2663188},
journal = {Ada Lett.},
month = oct,
pages = {103–104},
numpages = {2}
}

@article{Vehtari2016,
  title = {Practical Bayesian model evaluation using leave-one-out cross-validation and WAIC},
  volume = {27},
  ISSN = {1573-1375},
  url = {http://dx.doi.org/10.1007/s11222-016-9696-4},
  DOI = {10.1007/s11222-016-9696-4},
  number = {5},
  journal = {Statistics and Computing},
  publisher = {Springer Science and Business Media LLC},
  author = {Vehtari,  Aki and Gelman,  Andrew and Gabry,  Jonah},
  year = {2016},
  month = aug,
  pages = {1413–1432}
}

@article{Millidge2024,
  title = {Predictive coding networks for temporal prediction},
  volume = {20},
  ISSN = {1553-7358},
  url = {http://dx.doi.org/10.1371/journal.pcbi.1011183},
  DOI = {10.1371/journal.pcbi.1011183},
  number = {4},
  journal = {PLOS Computational Biology},
  publisher = {Public Library of Science (PLoS)},
  author = {Millidge,  Beren and Tang,  Mufeng and Osanlouy,  Mahyar and Harper,  Nicol S. and Bogacz,  Rafal},
  editor = {Latham,  Peter E.},
  year = {2024},
  month = apr,
  pages = {e1011183}
}

@ARTICLE{Deng2023,
  title        = "Causal reinforcement learning: A survey",
  author       = "Deng, Zhihong and Jiang, Jing and Long, Guodong and Zhang,
                  Chengqi",
  year         =  2023,
  primaryClass = "cs.LG",
  eprint       = "2307.01452"
}

@article{Binz2023,
  title = {Meta-learned models of cognition},
  volume = {47},
  ISSN = {1469-1825},
  url = {http://dx.doi.org/10.1017/S0140525X23003266},
  DOI = {10.1017/s0140525x23003266},
  journal = {Behavioral and Brain Sciences},
  publisher = {Cambridge University Press (CUP)},
  author = {Binz,  Marcel and Dasgupta,  Ishita and Jagadish,  Akshay K. and Botvinick,  Matthew and Wang,  Jane X. and Schulz,  Eric},
  year = {2023},
  month = nov 
}

@article{Chen2024,
  title = {The causal structure and computational value of narratives},
  volume = {28},
  ISSN = {1364-6613},
  url = {http://dx.doi.org/10.1016/j.tics.2024.04.003},
  DOI = {10.1016/j.tics.2024.04.003},
  number = {8},
  journal = {Trends in Cognitive Sciences},
  publisher = {Elsevier BV},
  author = {Chen,  Janice and Bornstein,  Aaron M.},
  year = {2024},
  month = aug,
  pages = {769–781}
}

@misc{willard:2023,
  doi = {10.48550/ARXIV.2307.09702},
  url = {https://arxiv.org/abs/2307.09702},
  author = {Willard,  Brandon T. and Louf,  Rémi},
  title = {Efficient Guided Generation for Large Language Models},
  publisher = {arXiv},
  year = {2023},
  copyright = {Creative Commons Attribution 4.0 International}
}

@article{Goddu2024,
  title = {The development of human causal learning and reasoning},
  volume = {3},
  ISSN = {2731-0574},
  url = {http://dx.doi.org/10.1038/s44159-024-00300-5},
  DOI = {10.1038/s44159-024-00300-5},
  number = {5},
  journal = {Nature Reviews Psychology},
  publisher = {Springer Science and Business Media LLC},
  author = {Goddu,  Mariel K. and Gopnik,  Alison},
  year = {2024},
  month = apr,
  pages = {319–339}
}
\end{document}